\documentclass[11pt]{article}

    \usepackage[preprint,nonatbib]{neurips_2023}

\usepackage{xcolor}
\usepackage{enumitem}
\usepackage{tikz}
\usetikzlibrary{positioning}

\usepackage[square,numbers]{natbib}
\usepackage[colorlinks]{hyperref}

\usepackage{url}

\newlist{checklist}{itemize}{1}
\setlist[checklist]{label={},leftmargin=!,before={\ttfamily}}

\title{Towards Regulatable AI Systems: \\
Technical Gaps and Policy Opportunities}

\author{%
  Xudong Shen\\
  National University of Singapore\\
  \texttt{xudong.shen@u.nus.edu} \\
  \And
  Hannah Brown \\
  National University of Singapore\\
  \texttt{hsbrown@comp.nus.edu.sg} \\
  \And
  Jiashu Tao \\
  National University of Singapore\\
  \texttt{taojiashu@u.nus.edu} \\
  \And
  Martin Strobel \\
  National University of Singapore\\
  \texttt{mstrobel@comp.nus.edu.sg} \\
  \And
  Yao Tong \\
  National University of Singapore\\
  \texttt{tongyao@u.nus.edu} \\
  \And
  Akshay Narayan \\
  National University of Singapore\\
  \texttt{akshay.narayan@nus.edu.sg} \\
  \AND
  Harold Soh \\
  National University of Singapore\\
  \texttt{harold@comp.nus.edu.sg} \\
  \AND
  Finale Doshi-Velez \\
  Harvard University\\
  \texttt{finale@seas.harvard.edu} \\
}

\begin{document}

\maketitle

\begin{abstract}
There is increasing attention being given to how to regulate AI systems.  As governing bodies grapple with what values to encapsulate into regulation, we consider the technical half of the question: To what extent can AI experts vet an AI system for adherence to regulatory requirements? We investigate this question through the lens of two public sector procurement checklists, identifying what we can do now, what should be possible with technical innovation, and what requirements need a more interdisciplinary approach.
\end{abstract}

\section{Introduction}

Regulations represent a fundamental mechanism through which governments fulfill their duty to protect citizens from a variety of threats including health risks, financial fraud, and discriminatory practices. As AI systems become more advanced and integrated into our lives, there has been a corresponding urgency to ensure they align with social values and norms, and that their benefits significantly outweigh any potential harm. In response to this imperative, legal and regulatory bodies globally are engaged in a concerted effort to develop comprehensive AI regulations~\cite{US_blueprint,EU_AI_act,China_genAI_draft}.

However, the increasing size, generality, opaqueness, and closed nature of present-day AI systems pose significant challenges to effective regulation~\cite{ziegler2019fine,bai2022constitutional}. Even when requirements can be articulated, it remains uncertain if and how we can verify an AI system's adherence to these standards. A requirement that cannot be checked will not effectively provide protection. If we believe that AI systems \emph{should} be regulated, then AI systems must be \emph{designed to be regulatable.}~\looseness=-1

In this article, we consider the following questions: \textbf{What innovations in AI systems are needed for them to be effectively regulated?  And in what areas will innovations in AI methods alone be insufficient, and more interdisciplinary approaches required?}

We explore the answer through the lens of public sector AI procurement checklists, which offer a pragmatic perspective on the broader challenges of regulatable AI systems. These checklists do more than outline criteria for government procurement; they represent concerted efforts to codify regulatory requirements for the adoption of AI systems in the public sector. This positions them uniquely at the intersection of policy-making and practical implementation. The desiderata distilled in these checklists are comprehensive and are reflected in still nascent efforts to regulate AI in the private sector~\cite{EU_AI_act,biden2023executive}. Furthermore, public sector procurement checklists are among the more developed AI regulations, some having gone through several rounds of refinement~\cite{2023AI}. We emphasize that we focus on public sector procurement checklists to make our discussion concrete; the technical innovations needed to satisfy those checklists are relevant to the wider discourse of creating regulatable AI systems.~\looseness=-1

Specifically, we closely examine the technical criteria from two existing procurement checklists: the World Economic Forum’s AI Procurement in a Box (WEF)~\cite{WEF} and the Canadian Directive on Automated Decision-Making (CDADM)~\cite{CDADM}. As illustrated in Figure~\ref{fig:mapping}, we first group the technical criteria contained in these two checklists into categories that will be familiar to AI researchers and engineers: (pre-training) data checks, (post-hoc) system monitoring, global explanation, local explanation, objective design, privacy, and human + AI systems. For each category, we briefly summarize existing technical approaches that could be used to construct AI systems that meet those criteria.  Next, we identify areas where relevant technical approaches may exist, but additional technical innovation is needed to be able to vet increasingly complex AI systems being used in increasingly varied contexts.  
For example, the proliferation of large language models comes with a significant difficulty in evaluation, due to factors such as their open-ended nature and data leakage. While innovative approaches like Holistic Evaluation of Language Models (HELM)~\cite{liang2022holistic} and Elo ratings~\cite{zheng2023judging} are proposed, accurately evaluating these models continues to be an unresolved issue that requires further technical advancements for effective regulation and oversight. Finally, we briefly outline aspects of these criteria that may seem technical but actually require interdisciplinary approaches to vet.

Throughout this exercise, we assume no concerns about expertise; that is, there are sufficiently qualified AI and domain experts to review whether the AI system meets the checklist criteria.  Our concern is to identify to what extent experts can currently vet AI systems against these criteria, and provide a (non-comprehensive, but concrete) list of directions for technical innovation to bridge the gap towards regulatable AI systems. If AI systems can be made verified against these checklists, then we will have made significant progress towards creating regulatable AI systems in general.~\looseness=-1

\section{Public Sector AI Procurement Checklists} 
\label{sec:AI_procurement_checklists}

\begin{figure}
    \centering
    \includegraphics[width=\textwidth]{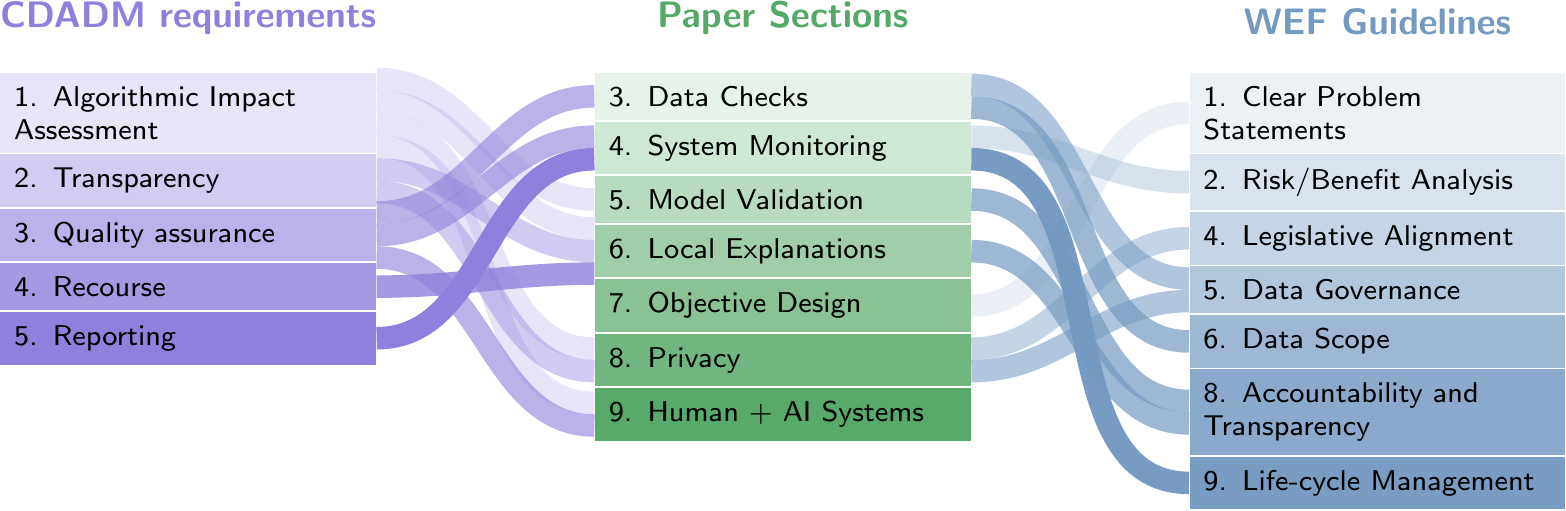}
    \caption{Connections between the main sections in this article and the high-level CDADM requirements and WEF guidelines. %
    We exclude the WEF guidelines on strategy alignment, multidisciplinary teams, and fair business competition which do not directly motivate CS research.}
    \label{fig:mapping}
\end{figure}

The public sector uses procurement checklists to ensure its purchased products align with its organizational needs and values.  In the context of AI procurement checklists, most pertinent to us are the items related to technical criteria, such as ensuring data quality, privacy, fairness, and appropriate monitoring and oversight. These technical criteria are also contained in broader but more nascent regulatory efforts.~\cite{EU_AI_act,biden2023executive}.

A reasonable question may be why focus on these particular checklists---after all, not all regulations are of the same quality.  We note both checklists have gone through extensive review. The directive on automated decision-making marks the Canadian government's first step toward managing the risks of AI in public administration. Initiated in 2016, this comprehensive effort involved a white paper, workshops, consultation sessions, and the formulation of working groups, drawing upon the expertise of scholars, civil society advocates, and governmental officials~\cite{Canada2024Responsible}. Officially enacted in April 2019, this directive has undergone three rounds of review~\cite{bitar20223rd} and was amended in 2021 and 2023 in response to these reviews. 
It established the groundwork for the broader 2022 Artificial Intelligence and Data Act (AIDA)~\cite{Canada2022Artificial} and a recent guide on the use of generative AI~\cite{Canada2024Guide}.~\looseness=-1

The "AI Procurement in a Box" by the World Economic Forum is a set of guidelines, with detailed examples, to assist governments in the creation of public sector procurement regulation.  It is the result of a 2018 "Unlocking Public Sector AI" initiative which included over 200 stakeholders across government, academia, and industry; the efficacy and applicability of the guidelines have been validated through two pilot studies conducted in the UK~\cite{World2020AI_UK} and Brazil~\cite{2022Unpacking}.~\looseness=-1

As noted above, both of the checklists we consider have gone through extensive review.  We also spoke to government officials who were using these checklists and carefully examined them with a red-teaming mindset. While no regulatory effort is perfect, we found that these checklists are fairly comprehensive, with perhaps the only technical area lacking being insufficient attention to HCI elements of how the AI would be integrated into its intended environment.  We found that their included technical criteria are relevant for broader efforts to create more regulatable AI systems.~\looseness=-1

For this paper, we aligned and grouped parts of the two checklists with existing AI research topics (outlined in Figure \ref{fig:mapping}). This process required combining information from different checklist sections, highlighting the differences in how AI researchers and policymakers approach the same problems.  For example, fairness is a large AI research topic, but interwoven throughout many points in CDADM and WEF, so it does not have its own section here.  This paper discusses all major sections of the CDADM but excludes some elements of security and non-expert training in the WEF due to space constraints.~\looseness=-1

\section{Inputs of the Model: (Pre-training) Data Checks} 
\label{sec:data_checks}

The characteristics of the training data have a large influence on the behavior of an AI system.  What checks must be done on these data before they are used to train models?  Motivations for the regulatory requirements in this section include data consent, data privacy (discussed in more detail in Section~\ref{sec:privacy}), and downstream impacts of data quality (e.g. on model performance, generalization and bias).  Examples of checklist criteria include: 

\begin{checklist}
\item CDADM 6.3.1: Before launching into production, developing processes so that the data and information used by the Automated Decision Systems are tested for unintended data biases and other factors that may unfairly impact the outcomes.

\item CDADM: 6.3.3: Validating that the data collected for, and used by, the Automated Decision System is relevant, accurate, up-to-date, and in accordance with the Policy on Service and Digital and the Privacy Act.

\item CDADM 6.3.4: Establishing measures to ensure that data used and generated by the automated decision system are traceable [fingerprinting], protected and accessed appropriately, and lawfully collected, used, retained, and disposed.

\item WEF: Assess whether relevant data will be available for the project […] Data is crucial for modern-day AI tools. You should determine, at a high level, data availability before starting your procurement process. This entails developing an understanding of what data might be required for the project.

\item WEF: Select data that fits criteria of fairness. For example, the data should be representative of the population that the AI solution will address, as well as being reasonably recent. 
\end{checklist}
 
\noindent The technical questions underlying these criteria have to do with data documentation procedures and checks that can expose potential risks in areas such as fairness, generalization, and privacy.

\subsection{What we know how to do} 

We have proxies for checking many properties in these criteria (data privacy, label quality, feature selection, fairness, etc.) using exploratory data analysis~\cite{tukey1977exploratory}. For example, we can inspect the annotation process and check inter-annotator agreement to get an idea of label quality~\cite{bayerl2011determines,paun2022statistical}. We can also measure (and correct for) imbalance in data if we are given group labels that segment the dataset~\cite{9423296,buolamwini2018gender}. We have techniques for identifying influential points~\cite{belsley2005regression}, outliers~\cite{ben2005outlier}, and mislabelled points~\cite{brodley1999identifying,pleiss2020identifying} which may cause models to exhibit poor performance or bias~\cite{field2012discovering}.

Further, there exist several standards for reporting dataset information~\cite{gebru2021datasheets, holland2020dataset,bender2018data,mcmillan2023data, pushkarna2022data}, including on the data curation process, that are designed to help expose potential biases and limitations for which the data may be used.  Sufficiently comprehensive data documentation facilitates investigation by both experts and the public.  In the realm of consent, non-consenting (opt-out) data checks\footnote{For example, artists can opt-out their work with Spawning (\url{https://spawning.ai}), which provides opt-out data checks as a service to AI system developers.} can give individuals control over how their data are used.

\subsection{Directions requiring additional AI innovation} 
\begin{itemize}
    \item \textbf{Metrics and Generalizability.}  
    More work is needed to connect the data metrics with impact on outcomes.  For example, we have reasonable tools to connect the uncertainty or measurement error in a distance sensor to effects on motion planning \cite{elbanhawi2014sampling}. However, if a traffic image dataset has a certain annotator disagreement score, what does that imply for an autonomous vehicle whose vision system is trained on those data? The question of generalizability also arises for data without explicit human annotation, such as internet-crawled language and vision datasets~\cite{2019t5,schuhmann2022laion}. In this case, what data checks can we perform to ensure that it will be appropriate for the domains in which the model is deployed? Data checks might lose their validity if the data is used outside its envisioned context.

    A particularly important category are metrics that capture similarities of different applications and thus capture scenarios in which a data set collected for one purpose may be used for another.  While this inquiry has received considerable attention in domain adaptation research~\cite{cortes2011domain, achille2019task2vec}, the data-centric perspective remains relatively unexplored~\cite{alvarez2020geometric}.  For example, a dataset collected for autonomous vehicles in one city \emph{might} be suitable in similar cities. But what statistics or meta-data would we need to be confident?  To ensure reliable utilization of datasets, additional metrics are necessary to precisely determine the range of applications for which a dataset can be safely used.

    \item \textbf{Data Quality Checks in the Context of Pretrained Models.} 
    Given the prevalence of large pre-trained models \cite{han2021pre} and (currently) limited transparency about their training data \cite{carlini2021extracting}, can we develop data checks that rely solely on accessing the model \cite{mitchell2019model}, or do certain types of checks require disclosure of specific information about the training data? Do checks for fine-tuning data---e.g., the traffic images used to tune an autonomous vehicle's vision system on top of an existing image classifier---differ from checks for pre-training data?
    
    \item \textbf{Unstructured Data.} For structured data, it is relatively easy to report statistics across features. For unstructured data like images or social media messages, existing standards focus on reporting the statistics of the meta-data \cite{gebru2021datasheets, holland2020dataset,bender2018data,mcmillan2023data, pushkarna2022data}.  However, is providing transparency about the meta-data sufficient?  For example, in the above scenario with the traffic images, is it sufficient to provide information e.g. about where the images were collected and what kinds of cameras were used?  Or might it be important to report certain information derived from the pixel values as well?  Similarly, if one had a collection of social media posts, would be it important to report certain information derived from the actual content, in addition to meta-data about the site and scraping procedure?
        
\end{itemize}

\subsection{Areas that require interdisciplinary engagement} 
The specific metrics that would enable meaningful inference about the quality of the data will depend on the application~\cite{heinrich2018requirements,kosmala2016assessing}.  Questions around bias and fairness are also inherently multi-faceted and will depend on the use-case~\cite{barocas2023fairness}.  Determining how data collection for AI systems respects copyright, obtains proper consent (opt-in versus opt-out), and avoids misrepresentation or detriment to the owner's benefits necessitates input from disciplines such as law, policy, and social science~\cite{kosinski2015facebook,de2023opt,jiang2023ai,Hao2022new}. Privacy tensions---what data is retained, what statistics are made public, what kind of access is granted to trusted auditors---must also be resolved within the broader socio-technical context~\cite{dick2023confidence}.

Furthermore, there is danger in living exclusively inside the data; cross-talks inside and outside of the data are necessary to detect many normative pitfalls. For example, bias can be introduced via the choices of labels (e.g. are non-binary labels included when labeling gender?) and the labeling process (e.g. whose perspective was being taken when an input was labeled as acceptable or problematic content?). Healthcare algorithms that demonstrate unbiased predictions of healthcare costs, but then use that prediction as a proxy for illness severity, may introduce bias because unequal access to care leads to lower healthcare spending by minority groups~\cite{obermeyer2019dissecting}. Detecting and addressing such issues in data necessitates active dialogue between the data realm and external perspectives. Section~\ref{sec:objective_design} delves deeper into the discussion of label choice concerns.

\section{Outputs of the Model: (Post-hoc) System Monitoring}
\label{sec:system_monitoring}

Once a system is deployed, it is essential to monitor its operations. These criteria have to do with monitoring for adverse outcomes and identifying unintended consequences, making that information available for scrutiny, and establishing contingencies if the system is behaving poorly.  Metrics to monitor the operations of a system also relate to methods for checking an AI's system's performance after it has been trained. Examples of checklist criteria include: 

\begin{checklist}
\item CDADM 6.3.2: Developing processes to monitor the outcomes of Automated Decision Systems to safeguard against unintentional outcomes and to verify compliance with institutional and program legislation, as well as this Directive, on a scheduled basis.

\item CDADM 6.3.6: Establishing contingency systems and/or processes as per Appendix C. (Which says: Ensure that contingency plans and/or backup systems are available should the Automated Decision System be unavailable.)

\item CDADM 6.5.1: Publishing information on the effectiveness and efficiency of the Automated Decision Systems in meeting program objectives on a website or service designated by the Treasury Board of Canada.

\item WEF: [T]here should be systematic and continuous risk monitoring during every stage of the AI solution’s life cycle, from design to post-implementation maintenance.

\item WEF: Testing the model on an ongoing basis is necessary to maintain its accuracy. An inaccurate model can result in erroneous decisions and affect users of public services.

\item WEF: Enable end-to-end auditability with a process log that gathers the data across the modelling, training, testing, verifying and implementation phases of the project life cycle. Such a log will allow for the variable accessibility and presentation of information with different users in mind to achieve interpretable and justifiable AI.
\end{checklist}

\noindent The technical questions associated with these criteria have to do with how to monitor performance and identify various kinds of drift and unusual results that warrant attention.

\subsection{What we know how to do} 
Given a specific metric, it is relatively easy to put monitoring into place.  We can easily check to ensure that the outputs of an AI do not exceed threshold values. Methods exist that establish distributions for "normal operation" and flag anomalous values during actual operation~\cite{gama2014survey}. These techniques can be employed to detect shifts in inputs and outputs, in model confidences and calibrations~\cite{barda2020addressing}, in derived quantities such as the top features used to make a prediction (allowing a person to check if a shift is sensible) and fairness metrics~\cite{ghosh2022algorithmic,barda2020addressing,9706306}.
We can learn a trend in how a particular quantity changes and see if that trend holds and whether any external shock occurs. In RL settings, we can monitor differences between expected and actual reward distributions. If the causal structure of the environment is known, monitoring checks can specifically identify new confounders and mediators. That said, all anomaly detection methods require some specification of what kinds of behavior represent a change or anomaly.  They may not capture every unintended consequence, and given sets of monitoring metrics may be gamed by an adversary.

More generally, we already have a set of norms around what kinds of tests should be run prior to an AI system being deployed (e.g. \cite{kreuzberger2023machine,smith2022artificial}).  AI developers should strive to test their systems with multiple independent, external datasets to ensure that their results are replicable (and be transparent if this kind of generalization has not been tested).  These datasets should include sufficient numbers of hard cases in their test sets, and results should be presented stratified by difficulty.  Similarly, one should provide stratified results on performance of cases similar and dissimilar to the training set.  Performance measures should be reported with respect to the real population proportions of each class, stratified by class, or be independent of base rates so that they can be correctly applied to the intended use-case and not the proportions present in the training set.

\subsection{Directions requiring additional AI innovation} 
\begin{itemize}
\item \textbf{Monitoring Many Metrics.}  Monitoring multiple metrics increases the risk of false positives and false negatives, which can overwhelm engineers. How can we monitor many metrics efficiently while not incorrectly flagging too many cases for review and not missing important deviations?  Relatedly, once in operation, what data should be gathered so that we can check additional metrics in the future? For example, while we can monitor fairness for known minority groups, what data should be logged during operation so that we can audit fairness when an unknown demographic group (e.g., an intersection of some legally protected attributes) contest for unfair outcomes~\cite{pmlr-v80-kearns18a}?  The question of what logs to retain only becomes more difficult when there are multiple AI systems interacting at fast rates, such as the many AI components operating within an autonomous vehicle.  These questions remain despite advances in MLOps \cite{kreuzberger2023machine}.

\item \textbf{Certification of Use Cases.} Across the very broad range of AI systems and contexts, can we certify the settings in which an AI system is supposed to work well? Can we assign a label to an AI model so that it is restricted to or from being applied to specific use cases? Consider, for example, the need to establish safeguards that prevent an open-access drug discovery model from being utilized for de novo design of biochemical weapons~\cite{urbina2022dual}.  Similarly, image generative models should be restricted from generating pornographic content.  Relatedly, can we provide confidence about the post-hoc performance of a deployed system on certified tasks while preventing a deployed system from being misused?
  
In formal verification, one mathematically checks that the formal model of a given system satisfies a desired property.  Formal verification is widely in safety-critical systems.  As AI systems enter safety-critical settings---such as autonomous driving or robot-assisted surgeries---it is essential that strong safety guarantees can be maintained.  Certifying neural networks for safety-critical systems is an active research area~\cite{sommer2020towards,singh2019abstract,baluta2021scalable,Katz2017ReluplexAE,Gehr2018AI2SA,xiao2022self}.
There are also early proposals to define standards for levels of AI system certification \cite{yap2021towards} (analogous to security standards~\cite{security_standards}) that have yet to be refined and adopted.

\item \textbf{Correcting Models after Deployment.} There exists some work on correcting deployed models in a way that does not require re-training end-to-end (e.g. unlearning~\cite{gupta2021adaptive,tarun2023fast,krishna2023towards,chourasia2023forget}, fine-tuning~\cite{hulora}, and in-context learning~\cite{weichain,chan2022data}).  But more work remains to be done, especially for AI systems with many interacting parts.

\item \textbf{Identifying Relevant Distribution Shift.} There are many possible types of shift: in input distributions, in the relationship between inputs and outputs, in the rewards (objective)---and these shifts can take many forms and occur in many ways.  For example, the acceleration of newer cars may be different, as well as what colors are popular.  Can we distinguish between relevant and irrelevant shifts (e.g., along the lines of \cite{chuang2020estimating})?  If the shifts happen in some uninterpretable embedding space, how can we explain them?

\item \textbf{Monitoring Agents that are Learning Online.} We can monitor for major adverse effects.  However, can we identify more subtle issues, such as initial signs of catastrophic forgetting, cheating, and other harms that occur while the agent continues to perform well on its reward metric? For instance, it would be advantageous to detect early signs of reckless or inappropriate driving behavior---such as reducing distances between the vehicle and pedestrians, or increased use of residential streets where children may be playing---in autonomous driving agents before any traffic accidents occur. Our understanding of unintended consequences continues to grow \cite{suresh2019framework,cabitza2017unintended} but the problem remains unsolved.

\end{itemize}

\subsection{Areas that require interdisciplinary engagement} 
At a high level, there will always need to be some kind of decision made about what needs to be monitored or prioritized in a given setting. There will also need to be decisions made about what kinds of safety promises or guarantees are needed e.g. how much shift is considered safe and acceptable, and how much is not, for example in healthcare~\cite{feng2022clinical}.  It is crucial to translate the monitored metrics into meaningful implications that enable people to make informed decisions within the broader socio-technical system~\cite{ortiz2023artificial}. For instance, in autonomous driving, comparing monitored metrics against human performance can inform decisions regarding human intervention.  Finally, the task of contingency planning for back-ups when models express unexpected or unwanted behaviors also requires an understanding of the broader socio-technical system.

\section{Inspecting the Model: Global Explanations for Model Validation}
\label{sec:global_explanation}
Global explanations describe a model as a whole and are often useful for inspection or oversight. The goal is to expose information about the model that would allow a domain expert to infer the existence of some kind of unobserved confounder, something about the model that is non-causal, and other limits on the scope of the model’s applicability.  Criteria related to global explanations include: 

\begin{checklist}
\item CDADM App. C: Plain language notice through all service delivery channels in use (Internet, in person, mail or telephone). In addition, publish documentation on relevant websites about the automated decision system, in plain language, describing: How the components work; 

\item  WEF: Public institutions cannot rely on black-box algorithms to justify decisions that affect individual and collective citizens’ rights, especially with the increased understanding about algorithmic bias and its discriminatory effects on access to public resources. There will be different considerations depending on the use case and application of AI that you are aiming to acquire, and you should plan to work with the supplier to explain the application for external scrutiny, ensuring your approach can be held to account. These considerations should link to the risk and impact assessment described in Guideline 2. Under certain scenarios, you could consider making it a requirement for providers to allow independent audit(s) of their solutions. This can help prevent or mitigate unintended outcomes. 

\item  WEF: Ensure that AI decision-making is as transparent as possible. – Encourage transparency of AI decision-making (i.e. the decisions and/or insights generated by AI). One way to do this is to encourage the use of explainable AI. You can also make it a requirement for the bidder to provide the required training and knowledge transfer to your team, even making your team part of the AI-implementation journey. Finally, you can ask for documentation that provides information about the algorithm (e.g. data used for training, whether the model is based on supervised, unsupervised or reinforcement learning, or any known biases).
\end{checklist}

\noindent Technical approaches associated with these criteria include the creation of small, inherently interpretable models with high performance, sharing certain parts or properties of a large model, and open-sourcing the model's code.

\subsection{What we know how to do} 
We can build inherently interpretable models (e.g. generalized additive models, decision trees, rule-based models, etc.) for tabular and other simple, relatively structured data \cite{rudin2019stop}. We have some tools for interpreting neural networks in terms of human-understandable components~\cite{raukur2022toward,nanda2023progress, olsson2022context}, such as circuits~\cite{wanginterpretability} or even natural language~\cite{bills2023language}. When possible, these tools provide a systematic approach to explain how tasks are performed in ML models in a human understandable way. Finally, we can partially explain neural networks and other complex models via methods such as distillation \cite{tan2018distill}, feature importance \cite{NIPS2017_7062}, or computing concept activation vectors \cite{samek2019towards}.

\subsection{Directions requiring additional AI innovation} 
\begin{itemize}
    \item \textbf{Inherently Interpretable Models for More Data Types.}  While some initial work exists for building inherently interpretable models for non-tabular data (e.g. for images or audio) \cite{chen2019looks}, this area is still nascent. Concept learning \cite{koh2020concept} on top of the input may be a useful strategy.

\item \textbf{Interactive “Openboxing” of Large Models.}  Can we build interactive, hierarchical, and semantically-aligned views of large models such that these models are (to some extent) inherently interpretable?   For example, a traffic image classifier that recognizes objects by multiplying object templates with transformation matrices~\cite{xu2021unsupervised}  would be more inherently explainable than another model without this hierarchical structure. Further, can we allow users to explore such explanations at different levels of fidelity for different contexts? As noted above, methods to extract information from larger models such as large language models exist (e.g., \cite{samek2019towards, molnar2020interpretable}) but have limitations with ways for people to effectively explore and understand larger models.  More work along the lines of \cite{bau2020understanding,strobelt2017lstmvis} is needed.

\item \textbf{Checking Value Alignment.}  
Whether it is criminal justice, benefits allocations, or autonomous driving, AI systems are increasingly used in situations that require value judgments.
How do we elicit and encode societal and individual values in diverse situations? What metrics can effectively measure value alignment?  How do we make this mapping transparent for others to understand the value choices made (e.g., the drivers of other cars next to the autonomous vehicle)?  Advancing exisiting work e.g., \cite{brown2021value,eckersley2018impossibility} is needed for our increasing use cases.

\end{itemize}

\subsection{Areas that require interdisciplinary engagement} 
There is a question of what to offer and to whom. For example, releasing the code and environment may allow some people to directly answer their questions~\cite{pineau2021improving}. Providing an explanation broadens who can inspect the model, including users and domain experts; however, what information to release, how it should be extracted, and how often during the life cycle of the model that information should be updated will depend on the use context.  We will also need mechanisms for people to request more information about a model as new concerns become apparent~\cite{Huq2019ART}.  Finally, all information release must be balanced with concerns about privacy and trade secrets.

\section{Inspecting the Model: Local Explanations about Individual Decisions}
\label{sec:local_explanation}

These criteria have to do with providing information to a user about a specific decision that is made, such as benefits denial.   In some cases, it may be sufficient to simply provide the information and logic that led to the decision (a meaningful explanation).  In other cases, it may be preferable to provide actionable ways to change the decision (recourse)~\citep{wachter2017counterfactual, karimi2021algorithmic}.  In the following, we use the term \emph{local explanation} to refer to explanations that are meant to provide insight about a particular decision, rather than about the model overall~\citep{mittelstadt2019explaining}. We use the term \emph{recourse} to refer a modification of the input that results in the output changing to the desired value.  

\begin{checklist}
\item CDADM 6.2.3: Providing a meaningful explanation to affected individuals of how and why the decision was made as prescribed in Appendix C. 

\item  CDADM 6.4.1: Providing clients with any applicable recourse options that are available to them to challenge the administrative decision. 

\item  CDADM App. C: In addition to any applicable legal requirement, ensuring that a meaningful explanation is provided with any decision that resulted in the denial of a benefit, a service, or other regulatory action. 

\item  WEF: Explore mechanisms to enable interpretability of the algorithms internally and externally as a means of establishing accountability and contestability. – With AI solutions that make decisions affecting people’s rights and benefits, it is less important to know exactly how a machine-learning model has arrived at a result if we can show logical steps to achieving the outcome. In other words, the ability to know how and why a model performed in the way it did is a more appropriate means of evaluating transparency in the context of AI. For example, this might include what training data was used, which variables have contributed most to a result, and the types of audit and assurance the model went through in relation to systemic issues such as discrimination and fairness. This should be set out as documentation needed by your supplier. – It is also important to consider the potential tension between explainability and accuracy of AI when acquiring AI solutions. Classic statistical techniques such as decision-tree models are easier to explain but might have less predictive power, whereas more complex models,such as neural networks, have high predictive power but are considered to be black boxes.
\end{checklist}

\noindent Approaches for creating local explanations rely heavily on a notion of local region, and thus some notion of distance.  Some inputs are more easily explained than others, and any explanation can introduce privacy risks.

\subsection{What we know how to do} 
There are many techniques of providing local explanations for a model \cite{doshi2017towards,lakkaraju2016interpretable,ribeiro2016should,ribeiro2018anchors,lundberg2017unified, simonyan2013deep}. Specifically, given a definition of distance, we can find a counterfactual: the closest point such that the model’s output is a desired class~\citep{guidotti2022counterfactual, wachter2017counterfactual}.  This can be used to help an individual determine what features set them apart compared to a nearby alternative, and also set the foundation for recourse (if those features can be changed) \cite{karimi2021algorithmic}.

\subsection{Directions requiring additional AI innovation}
\begin{itemize}
    \item \textbf{Defining Distance Metrics.}  As noted above, local explanations rely heavily on notions of nearby data.  It can be difficult to adjudicate what correlations in the data should be preserved and what should not. For example, if there are correlations between the kind of sign and the geographic location in a traffic image data set, should those correlations be retained in the distance metric? What about for race and postal codes or sex and hormone levels? Some work exists on using human input to define the appropriate distance metric for the purposes of explanation and recourse \cite{karimi2021algorithmic}, but more is needed.

    \item \textbf{Data without Interpretable Dimensions.}  The challenges associated with choosing distance metrics are exacerbated when the individual dimensions of the data are not interpretable.  For example, suppose we have a medical imaging task in which the AI system claims that certain cells represent a certain type of cancer, or a face recognition task in which the AI system claims that the face in a security video matches a face in a government database.  What is a meaningful explanation~\citep{mertes2022ganterfactual} in this case? Does it take the form of other images in the dataset (which may create privacy issues)? Should it involve first summarizing the input into interpretable concepts~\citep{kim2018interpretability, ghorbani2019towards}?  Similar issues arise with text~\citep{wu2021polyjuice} and timeseries data~\citep{ates2021counterfactual}.
    
    \item \textbf{Provenance Adjudication.}  We may want to know if particular training datum was used in a particular way to generate the given output. For example, we may be curious if a traffic sign mis-classificatoin could be attributed to a specific mislabel example, or we may need to resolve copyright issues from AI-generated text and images. This is possible in small models, but in very nascent stages for large models (e.g., LLMs~\cite{vyas2023provable} and diffusion-based image generation models~\cite{dai2023training}).
    
    \item \textbf{Handling Out of Distribution Data.}  The idea behind recourse is that it gives a person a path toward getting the outcome they desire.  For example, if a loan applicant is told that paying off their debts would make them eligible for the loan, then they would expect to get the loan once the debts are paid.  However, if the applicant's data is very far from the training data, then the AI-produced recourse may indeed change the model's output, but would not be accepted by the loan officer in a real context.

    \item \textbf{Tradeoffs between Explainability and Privacy/Security.} Releasing information for auditing or recourse may allow bad actors access to private information \cite{shokri2021privacy} or to game the system \cite{pawelczyk2023privacy}.  For example, explanations in the form of training samples, like those of the traffic images, may allow actors to learn not only how to trick the autonomous vehicle, but also learn about other elements of those images (that are not road signs). Advancing existing work e.g. \cite{tsirtsis2020decisions} is necessary to understand the resulting dynamics.  
\end{itemize}

\subsection{Areas that require interdisciplinary engagement} 
The biggest question raised by these guidelines is what is the definition of a “meaningful explanation"~\citep{sosa1997meaningful}. This definitions will depend on the socio-technical context of the task---contesting a loan denial, a medical error, or a benefits denial may require different kinds of explanations.  Different kinds of users may also require different explanations.

Relatedly, the purpose of the information provided for recourse will vary across contexts. For one task, it may be enough to provide only one recourse, while for others it may be necessary to provide multiple options~\cite{teso2019explanatory,sokol2020one}. In other contexts, the user might benefit from an interactive system to explore different options.  For example, they could themselves wish to navigate changes and see if they would result in a favorable loan decision.

Finally, it may be that a recourse generated from a local explanation may not be the appropriate way to assist a user unhappy with a decision~\cite{alvarez2018robustness,adebayo2018sanity}.  For example, suppose someone is convinced that a voice-based covid test is in error about their disease status. Rather than providing an explanation of the voice features used to make the decision, the appropriate recourse may be to allow that person to take a traditional covid test instead.  We also note that certain situations may require a justification (rationale for why a decision is right with respect to laws, norms, and other aspects of the context) rather than explanation (what features the AI used to generate the output).

\section{Designing the Model: Objective Design}
\label{sec:objective_design}

All AI systems require formulating goals in precise, mathematical terms.  Objective design converts general goals (e.g. drive safely) into precise mathematical terms~\cite{bernardi2019150,hennig2007some}.  This distillation process is fraught with potential pitfalls; an incorrect conversation will result in the AI behaving in unintended ways.  For example, encoding safe driving as always ceding the right of way may result in an autonomous vehicle that never makes a turn at a busy intersection. Collaboration with stakeholders during the objective design process can help ensure the true goals are addressed, rather than a proxy that may not result in the desired behavior. Documentation of the objective design process must be sufficiently transparent to ensure calibrated trust from stakeholders. Examples of criteria include:

\begin{checklist}
\item WEF: Focus on developing a clear problem statement, rather than on detailing the specifications of a solution. - AI technologies are developing rapidly, with new technologies and products constantly being introduced to the market. By focusing on describing the challenges and/ or opportunities that you want to address and drawing on the expertise of technology partners, you can better decipher what technology is most appropriate for the issue at hand. By focusing on the challenge and/or opportunity, you might also discover a higher-priority issue, or realize you were focusing on a symptom rather than the root cause.
\end{checklist}

\noindent The criteria above encourage public servants to identify their actual goals and then allow the engineers to deliver.  To be able to deliver, however, the AI engineers must be able to convert the problem statement into precise terms.

\subsection{What we know how to do} 
In some cases, it is possible to decompose a complex task into simpler components.  For example, in the context of an autonomous vehicle, we might evaluate a perception system for its ability to identify and forecast the trajectories of other objects in its environment, and the ability of a planner to make safe decisions given this information.  Algorithms for multi-objective optimization can find a Pareto front of options corresponding to different trade-offs between desiderata \cite{sawaragi1985theory,soh2011evolving}. There is also recent work in inferring what objectives are truly desired given observed reward functions~\cite{hadfield2017inverse}.

\subsection{Directions requiring additional AI innovation} 
\begin{itemize}
\item \textbf{Metrics for Metrics: Measuring Match to Goals.} 
  What are the measures that can be used to determine whether some technical objective matches our policy goals?  Objective and reward design are relatively well-studied in some domains, such as reinforcement learning~\cite{singh2009rewards,hadfield2017inverse}, but unsolved for the many more situations---from autonomous vehicles to email text completion---in which we see AI systems used today.  Further, our goals may be multi-faceted; the objective must not only be faithful to our goal but also transparent in how it is faithful.

\item \textbf{Properties of Popular Objective Functions.}  There are many objective functions used for their computational convenience and statistical properties  (squared loss, log likelihood, etc.).  Because they are so popular, their statistical properties under various conditions are often well-understood~\cite{shalev2014understanding}.  For example, we may know that L1 losses are more robust than L2; we may know that decreased model capacity (e.g. fitting a line) can make a model more prone to being swayed by influential points.  However, how do these very technical understandings of statistical properties relate to more complex goals, including reward hacking and other short-cut risks? Better understanding of these properties could enable better matching between popular losses and broader policy goals.

\item \textbf{Robustness to a Variety of Objectives.} 
Further research is necessary to create agents that excel across a range of objectives. In RL, this research can strengthen the robustness of learned policies when objectives are not perfectly specified~\cite{moos2022robust, pinto2017robust}. It also applies to language models, which are trained with the next token prediction task but asked to perform agentic tasks with various constraints and objectives~\cite{andreas-2022-language}, and in different worlds~\cite{lecun2022path}.

\item \textbf{Computational Constraints for More Robust Objectives.} Related to the above, there are a variety of computational constraints and regularizers that often make objectives more robust to imperfect specifications. 
These include encouraging smoothness (e.g. Lipschitzness), sparsity, and robustness to certain types of uncertainties (e.g. \cite{eckersley2018impossibility}, and distributionally robust optimization~\cite{rahimian2019distributionally}.
).  However, work remains to be done to more strongly connect what these computational tools do in the context of aligning the technical formulation with the true goal.

Furthermore, some constraints and regularizations are difficult to express and/or operationalize in analytical forms; instead, they are incorporated directly into the training procedure, such as adversarial training~\cite{tramer2018ensemble}.  
Relatedly, additional work is needed to effectively optimize objectives with multiple criteria---whether those are constraints, regularizers, or competing terms: Simply writing down an objective does not make it easy to optimize. As additional terms are added to the objective, the question of how to weigh them to achieve the desired behavior also becomes more complex.  

\item \textbf{Understanding Connections between Objectives and Learnt Model Behavior.} 
Can we efficiently explain how changes in the objective function impact model behavior?
Conversely, can we explain policies in terms of compatible reward functions?  Can we efficiently identify where two reward functions may result in different policies in human-understandable terms?  Some prior works try to answer this \cite{gajcin2022contrastive}; however, more analysis will help refine the reward function to better match the intended objectives. One further question is disentangling which model behavior is the result of an objective and which is the result of training data. For example, the mix of possibly conflicting beliefs in a text corpus will influence how language models trained on it behave, though all have the same objective.~\cite{andreas-2022-language}

\item \textbf{Inferring Goals from Observed Behavior.}  In some cases, we may have examples of decisions or outputs that we know align with the true goal (e.g. safe driving trajectories).  However, the inverse problem of inferring rewards from behavior is not identifiable. Advancing techniques\cite{arora2021survey} to help disambiguate important elements of the reward function can help ensure that the learned policy aligns with the desired objectives, leading to improved performance and generalization.

\end{itemize}

\subsection{Areas that require interdisciplinary engagement} 
Creating goals at a policy level requires considering factors such as contextual relevance, attainability, and alignment with overarching desiderata~\cite{hulten2018building}. Ethical concerns associated with the power and impact of AI systems may also be taken into account. Moreover, sometimes the objective remains unclear even at the policy level, making it more difficult to design proper objectives for the AI systems much less validate and explain them. For example, while AI systems are commonly utilized in criminal justice~\cite{zilka2022transparency,kleinberg2018human}, there is often a lack of clarity regarding how they define and measure crime~\cite{isaac2017hope}, and the data may not accurately reflect the true objectives of judges~\cite{cilevics2020justice,richardson2021racial}.

\section{Designing the Model: Privacy}
\label{sec:privacy}
Bad actors may use transparency about the data, code, and model for identifying private information about individuals. There are a number of examples of regulatory criteria relating to privacy concerns, including:

\begin{checklist}
\item CDADM 6.2.6: Releasing custom source code owned by the Government of Canada as per the requirements specified in section A.2.3.8 of the Directive on… 

\item  CDADM App. C: Plain language notice through all service delivery channels in use (Internet, in person, mail, or telephone). In addition, publish documentation on relevant websites about the automated decision system, in plain language, describing: A description of the training data, or a link to the anonymized training data if this data is publicly available. 

\item WEF: There are many anonymization techniques to help safeguard data privacy, including data aggregation, masking, and synthetic data.  Keep in mind, however, that you must manage anonymized data as carefully as the original data, since it may inadvertently expose important insights. RFPs should encourage innovative technological approaches, such as those mentioned above, that make less intrusive use of data or that achieve the same or similar outcomes with less sensitive datasets. 

\item WEF: As important as data protection is, not all data is sensitive (e.g. open-government data is freely accessible online). All data, sensitive or not, must have its integrity safeguarded, but it is not necessary to keep non-sensitive data behind closed doors. It is important to assess the privacy needs of different datasets to determine the right level of protection. Normally, personally identifiable information (PII), such as financial and health data, is considered extremely sensitive. The RFP needs to reflect data governance requirements for both the procurement process and the project that are in accordance with the nature of the data.
\end{checklist}

\noindent However, the language in these regulations leaves a number of issues unspecified, including a standardized, meaningful definition for privacy, and assumes that we are currently able to properly assess the privacy of a dataset and anonymize data, which are currently open research questions.

\subsection{What we know how to do} 
Differential privacy is a widely-accepted theoretical notion of privacy \cite{wood2018differential, dwork2006differential}.  In settings where this notion of privacy is appropriate, we have differentially private algorithms that can calculate statistical properties of data~\cite{dwork2014algorithmic}, train machine learning models~\cite{abadi2016deep, papernot2016semi}, and generate synthetic data~\cite{cai2023privlava}.  Many other privacy notions exist~\cite{samarati1998protecting,machanavajjhala2007diversity,li2006t}.  Choosing which privacy notion to use in a particular setting remains an open question.

\subsection{Directions requiring additional AI innovation}
\begin{itemize}
\item \textbf{Better Tradeoffs between (differential) Privacy and (predictive) Performance.} In general, differentially-private models have lower predictive performance than models without privacy guarantees \cite{bagdasaryan2019differential}.  How can that gap be closed?  Related questions include: Can we ensure models are private even with many queries and in conjunction with public data? What can we maximally expose about a model and training data statistics in a way that is still private? Can we precisely state what cannot be exposed, e.g. a long tail has been left out \cite{feldman2020does}?  (Note: if we can make this precise, then certain information could be made public as it poses no privacy risk, and other information may be available only to a trusted auditor.)

\item \textbf{Creating and Assessing Privacy Definitions.} How can we define privacy appropriately and meaningfully for different types of data? (e.g. trajectories, text \cite{brown2022does}, etc.). What do current definitions of privacy actually achieve on these data?

\item \textbf{Privacy via Minimal Data Collection.}  Can we collect only the input information needed for each decision, which may involve collecting different inputs for different people~\cite{tran2023data}?  What privacy risks are mitigated by this approach?  Are new risks introduced because what inputs are measured is new information?

\item \textbf{Private Generative Models.} The main focus of existing work is on classification. So, there are many open questions when it comes to the privacy of generative models~\cite{carlini2023extracting,carlini2021extracting,carlini2022quantifying,ippolito2022preventing}. For example: How can we prevent a generative model from replicating training data? Is there a difference between a private generative model and adding noise to data? Is there a benefit to a private generative model vs. noised data? Are empirical methods to ensure privacy e.g. via reinforcement learning with human feedback~\cite{ziegler2019fine}, sufficient?

\item \textbf{Effective Machine Unlearning.}  In some cases, people may be allowed to elect to have the influence of their data removed after the model has been trained. Methods have been created to remove the influence of specific inputs from the data, but these are still in progress~\cite{krishna2023towards,chourasia2023forget}, especially for generative models~\cite{schramowski2022safe,gandikota2023erasing,heng2023selective}.
\end{itemize}

\subsection{Areas that require interdisciplinary engagement} 
Current private models still allow third parties to infer private information via access to additional, publicly available data. We need to develop new notions of privacy for this setting~\cite{brown2022does}. Broader discussion is also needed regarding what to do if privacy guarantees sacrifice predictive performance, especially if the sacrifice is primarily to underrepresented groups \cite{bagdasaryan2019differential}.  More generally, the appropriate definition of privacy, and how strict the privacy guarantee must be (e.g., via hyperparameter settings), will depend on the setting~\cite{dwork2019differential} and must be made transparent.  For example, claiming a model is differentially private when it has a very large epsilon may be misleading.  

Finally, while this section has focused on privacy, we also note that there are many security concerns must also be considered in a holistic manner. There are clear limitations to what can be achieved with respect to adversarial actors.  If training data are available, a state actor or a large industry actor could (re)create a model. Once a model or training technique is out, we really cannot control its use. Unlimited public access to a model (via queries) intrinsically allows an adversary to learn about the model and the training data.

\section{Interacting with the Model: Human + AI Systems}
\label{sec:humans_in_the_loop}

AI regulations frequently emphasize the involvement of humans in various stages of the decision-making process. Often the intent is for the human decision-maker to vet an AI recommendation, take responsibility for the final decision, and intervene in case of emergency situations and system failures. We also consider the case of learning from human input.  Examples of related criteria include: 

\begin{checklist}
\item CDADM App. C: Decisions cannot be made without having specific human intervention points during the decision-making process; and the final decision must be made by a human. 

\item CDADM 6.3.6: Establishing contingency systems and/or processes as per Appendix C. (Which says: Ensure that contingency plans and/or backup systems are available should the Automated Decision System be unavailable.)
\end{checklist}

\noindent Technical approaches associated with these criteria include combining information from multiple experts, as well as ways to ensure that humans are fully engaged in the decisions.

\subsection{What we know how to do} 
There has been significant work on learning from humans. We can apply methods such as imitation learning \cite{hussein2017imitation,mandlekar2020human} and reinforcement learning~\cite{macglashan2017interactive} from human feedback to orient the model based on expert control or learn human intentions/preferences~\cite{palan2019learning}. Active learning techniques can be used to proactively ask for information to improve a model from humans~\cite{ware2001interactive,ren2021survey}.  Finally, we also have methods for humans to take the initiative to correct an agent (e.g.,~\citep{ross2011reduction, madras2018predict,mozannar2020consistent}). While methods in uncertainty quantification are always being improved, for the purposes of flagging uncertain inputs for human inspection \cite{geifman2017selective}, our current methods are reasonable.

\subsection{Directions requiring additional AI innovation}
\begin{itemize}

\item \textbf{HCI Methods for Avoiding Cognitive Biases.} 
Humans have many cognitive biases and limitations. If a system behaves most of the time, people may start to over-rely on it.
Confirmation bias can accompany backward reasoning (people finding ways to justify a given decision) but can be mitigated if a person performs forward reasoning first (looking at the evidence) \cite{bondi2022role}. Bias can also come from imperfect information fusion, e.g., if a human inspects the input data and then views an AI prediction based on the same input data, they may falsely believe that the AI prediction is a new, independent piece of information.  For example, we may be concerned that a clinician forms an opinion from patient data, and then sees an AI opinion based on the same data, they may falsely treat the AI opinion a new, independent form of evidence. Appropriate human+AI interaction can help mitigate these biases. 

\item \textbf{Shared Mental Models and Semantic Alignment.}  Shared
  mental models---between the human and the AI system, between the AI system and the human---are essential for effective human+AI interaction~\citep{andrews2023shared}. While there exists work in which agents use or create models of humans (e.g.,~\citep{carroll19}) to facilitate interaction, including modeling a person's latent states such as cognitive workload and emotions (e.g.,~\citep{ong2019computational}), it remains an open question as to how to develop and validate these methods for increasing number of human+AI use cases.

    One particularly important area is semantic alignment between the way humans organize concepts and the way modern AI systems encode representations.  Grounding terms has a long history in AI~\cite{harnad1990symbol} and innovation is needed for our modern settings.

\item \textbf{Humans-in-the-Loop in Time-Constrained Settings.} How can we include humans in the loop when decisions have to be made quickly e.g. industrial robots in emergency scenarios involving human workers? It is crucial that automated systems can fail gracefully and hand-over control to humans, even in time-constrained settings~\cite{russell2016motor}. 

\item \textbf{Human+AI for Test-time Validation of Large Surface Models.} Models with large output surfaces (e.g. LLMs) will be difficult to evaluate via prospective metrics; we need methods for people to assist in their validation at the task-time (e.g. as proposed in \cite{doshivelez2023contextual}).

\item \textbf{Evaluation and Design of Realistic Human-in-the-Loop Systems.} Most current testing is for lay user and consumer applications, where risks and costs are minimal.  However, evaluation in other settings is more challenging: Integrating a new interactive system into an existing workflow may require not only significant software effort, but also training of users.  In high-stakes settings such as healthcare, criminal justice, and major financial decisions, there is a risk of real harm to people.  How can we evaluate and design for these cases? Building more general knowledge about human-in-the-loop systems and developing smarter experimental designs may help reduce these burdens. So might validating methods for piloting methods in offline or de-risked ways that may still inform the target application.  Relatedly, standard procedures are needed for evaluating and monitoring human-in-the-loop systems.

\end{itemize}

\subsection{Areas that require interdisciplinary engagement}
Shared human+AI decision-making is an interdisciplinary area involving social science, psychology, cognitive science, etc.,~\cite{cila2022designing}. Fortunately, HCI research already has connections to these fields~\cite{cross2021mind,srinivasan2021explanation,sundar2020rise}. Furthermore, the design adoption of new tools into workplaces is well-studied in design, human factors research, and management and operations science~\cite{moore2019osh,agrawal2019economics,tyler2023ai}---and require interdisciplinary teams with appropriate expertise.  These interdisciplinary efforts will help inform decisions about whether, how, and which humans to include in the loop, as well as how a system that is expecting human input should respond to inappropriate, slow, or absent input from the human. 

\section{Conclusion}
In this document, we examined the technical criteria in two real regulatory frameworks---the Canadian Directive on Automated Decision-Making and World Economic Forum AI Procurement in a Box.  We find that we only have some of the tools needed to ascertain whether an AI system meets the stated requirements.  We list several concrete directions for AI innovation that, if addressed, would improve our ability to create regulatable AI systems.  

\paragraph{Acknowledgements.} The authors thank Andrew Ross, Siddharth Swaroop, Rishav Chourasia, Himabindu Lakkaraju, and Brian Lim; all participants of NUS Responsible, Regulatable AI Working Group 2022-2023 including Limsoon Wong, Angela Yao, Suparna Ghanvatkar, and Davin Choo.

\bibliographystyle{abbrvnat}
\bibliography{main.bib}

\end{document}